# How Much Determinism Can Fine-Tuning Buy? Margin Regularisation Against Batch-Induced Token Flips

Charan Teja Nagisettygari
*School of Computer Science*
*Taylor's University*
Subang Jaya, Malaysia
0371148@sd.taylors.edu.my

Dr. Siva Raja Sindiramutty
*School of Computer Science*
*Taylor's University*
Subang Jaya, Malaysia
Siva.sindiramutty@taylors.edu.my

*Abstract*—Greedy decoding at zero temperature is widely assumed to be deterministic, although the exact same prompt may yield different tokens depending on the number of other prompts sharing the batch for inference. The currently available solutions operate during inference, trading determinism for throughput. Can we train a model in such a way that fewer decoding steps would be susceptible? We characterize the perturbation empirically across six model families and three numerical precisions and observe two empirical facts about the measurement: the perturbation depends exactly on the representable precision (8.00× for bfloat16 vs float16, 8192.0× for float16 vs float32, in line with predictions based on mantissa bits) and the batch invariance happens accidentally at certain tensor shapes, allowing a practitioner to achieve zero nondeterminism merely by setting the sequence length. Then we apply hinge penalization of the model's top-1/top-2 logit gap, scaled to twice the measured noise quantile. On GSM8K dataset it leads to a 35% reduction in token flips and divergence from 53.8% to 38.1%, at the cost of 1.3–2.9 accuracy points, increasing monotonically with the margin size. It does *not* affect the gap threshold sufficient for determinism: the logit gaps of surviving flips are unchanged (median 0.125 for all arms), as the hinge is a soft penalty and the largest hardware perturbation remains the same. We reject seven hypotheses, including a per-model viability diagnosis, holding on six models and falsified on the seventh.

## I. Introduction

Greedy decoding is taken as a reproducible primitive in the wild. The evaluation setup, LLM-as-judge workflow, the regression suite, and distillation loop all rely on the assumption that a computation is uniquely determined by its input prompt and decoding strategy. In practice, however, this assumption is not true: the same prompt may produce distinct token sequences depending on whether it is decoded as part of a batch or alone, even when the temperature is set to zero [1]– [3]. This phenomenon was observed both in production and offline setting, with temperature zero variation being present in both raw output and parsed answer [4] as well as code generation task [5]. And the effect is big enough to invalidate the whole benchmark evaluation protocol without taking it into account [6].

The reason is not stochastic, but numerical. Floating-point additions are not associative [7], [8], and high-throughput kernels pick summation order depending on the shape of the operator. Consequently, requests that share a batch are reduced differently than those that do not, producing last bit differences in the log-probability vector that may flip the argmax when two top scores are comparable.

### A. Existing remedies and their cost

There are two categories of solutions, and they are all applied during inference. Batch invariant operators and upcast layers ensure that every reduction is applied using a unique policy [1], [2], ensuring determinism at the cost of 2x slowdown; previous work includes ordering reductions for various tensor parallelisms [9], and rescheduling deterministic attention to get back training throughput [10]. Selective verification keeps the fast path and recomputes only the steps judged at risk [3], [11], which on the models we replicate requires verifying between 8% and 18.6% of decoded steps. Both pay for determinism in serving cost, on every request, forever.

### B. This work

We ask a different question: can the *model* be changed so that fewer steps are at risk in the first place? The lever is the top-1/top-2 logit gap, which prior work has already shown to be the operative risk signal [3]. A training-time intervention that widens this gap costs nothing at serving.

We report three kinds of contribution.

Measurement. We characterise the perturbation across six model families, three precisions and two GPU architectures. Two results are new. The perturbation scales *exactly* with the representable precision of the number format, at ratios matching mantissa-bit predictions to the digits we report (Section III-D). And batch invariance arises *accidentally* at particular tensor shapes on CUDA, such that the same model shows ≈ 55% perturbed steps at one sequence length and exactly zero at another (Section III-E). Any reported flip rate is therefore meaningless without its sequence length and batch size.

Method. We fine-tune with a hinge penalty on the model's own top-1/top-2 gap, with the margin set to twice the measured noise quantile rather than tuned. Across six models in four families, efficiency peaks at this derived value.

Negative results. We report, with the evidence that killed them, seven hypotheses that did not survive including a permodel viability diagnostic that predicted six outcomes correctly under pre-registration ($p \approx 0.002$) and was then refuted by two models sharing an identical diagnostic value with opposite outcomes. Most importantly, we show that margin training reduces how *often* flips occur but not the worst-case gap at which one can occur, and give the mechanism for why.

## II. Background and Related Work

### A. Why batching changes the output

For a protected request $q$, let $\ell_q \in \mathrm{R}^{|V|}$ be the logit vector produced by the default forward pass at a decode step, and let $\ell_q^{(1)}$, $\ell_q^{(2)}$ denote its two largest entries. We write the *gap*

$$g_t = \ell_{q,t}^{(1)} - \ell_{q,t}^{(2)}. \tag{1}$$

Running $q$ inside a batch of size $N$ rather than alone changes the shape of every matrix multiplication in the network. Kernel libraries choose a reduction strategy from that shape, so partial sums accumulate in a different order and the logits differ in their last representable bits. We denote the resulting perturbation to the gap

$$\delta_t = g_t^{(N)} - g_t^{(1)}. \tag{2}$$

A *token flip* occurs when $g_t^{(N)} < 0$ , that is, when the perturbation is large enough to reverse the ordering of the two leading candidates. Note that $\delta_t$ is not random: for a fixed shape the kernel is run-to-run deterministic [1], so the perturbation is better described as a small set of arithmetic paths than as noise. Yashwanth does the same from an arithmetic standpoint, arguing that the error associated with batch-invariant matrix multiplication is highly structured, not random across the output [12]. We keep using the distribution terminology, since the route followed cannot be controlled from the request's point of view.

### B. Inference-time determinism

Batch-invariant operators from Thinking Machines Lab [1] and LayerCast [2] modify either reductions or linear layers such that the outcome is independent of the batch dimensions. The fast pass is preserved in LLM-42 [11], while each decoded token is checked against a deterministic recomputation. MarginGate [3] narrows this: it triggers the verifier only when $g_t < \tau$, restoring sequence-level determinism on Llama-3.18B while verifying 18.56% of synchronous decode steps. Determinism has also been pursued by fixing reduction order across tensor-parallel configurations [9] and by rescheduling deterministic attention kernels [10]. All of these act on execution rather than on weights.

MarginGate also supplies the evaluation protocol we adopt. Their reported flip rates across five models and three benchmarks span 0.3–1.3% of synchronous decode steps, and they establish that the cross-batch alternative token is almost always among the top few candidates, which is what makes the top1/top-2 gap a deployable risk signal.

### C. Training-time margin objectives

Margin-augmented cross-entropy is not new. MCEL [13] connects bit-error tolerance in accelerators to output-layer classification margins and proposes a margin term with an interpretable parameter, arguing it outperforms training-time error injection. Analog foundation models [14] fine-tune LLMs against hardware noise at scale using injection plus distillation.

Margin-sensitive objectives are also established in LLM training more broadly, though at the level of whole responses rather than single-token logits: margin matching preference optimisation controls response-level reward separation [15].

What is new here, to our knowledge, is the empirical calibration of the margin from measured hardware noise, the application to batch-induced autoregressive nondeterminism rather than to bit errors or analog weight noise, and the characterisation results of Section III. We are not aware of prior work that applies a perturbation-calibrated token-margin objective to this failure mode, or that reports what such an objective can and cannot achieve.

Two cautions from adjacent literature inform the design. Hyperfitting demonstrates how optimization can result in an extremely low output entropy, enhancing greedy generation but reducing validation performance [16], while calibration studies show that instruction tuning is already biased towards overconfidence [17], [18]. A determinism objective should therefore enforce only the margin a measured perturbation requires, rather than maximising confidence globally.

## III. Characterising the Perturbation

### A. Protocol

Every measurement in this paper follows the same protocol. We take $N$ fixed-length token sequences and run each one alone, recording the top-5 logits and their token identifiers. We then run the identical sequences a second time alone, an *identity control* and finally inside a batch of size $B$, gathering logits at the token identifiers recorded in the first pass so that the same quantity is compared throughout.

The identity control is not optional. Without it, a measured difference cannot be attributed to batching rather than to runto-run variation. It returned exactly zero flips in every valid run reported in this paper.

### B. The phenomenon, and where the risk lives

On Qwen2.5-0.5B in float16 over 32,640 decision steps (Table I), batching from $B$=1 to $B$=32 produced 62 flips (0.19%) against an identity control of exactly zero. This is the same order as the 0.3–1.3% reported by prior work [3] on larger models.

TABLE I
Flip rate by logit gap. Qwen2.5-0.5B, float16, $B$=1 vs. $B$=32, 32,640 prefill decision steps. Identity control: 0 flips.

| Gap range | Steps | Flips | Rate |
|---|---|---|---|
| [0,0.05) | 1,405 | 62 | 4.41% |
| [0.05,0.1) | 1,158 | 0 | 0% |
| [0.1,0.25) | 3,139 | 0 | 0% |
| [0.25,0.5) | 4,682 | 0 | 0% |
| [0.5,1.0) | 6,720 | 0 | 0% |
| [1.0,2.0) | 7,370 | 0 | 0% |
| [2.0,∞) | 8,166 | 0 | 0% |

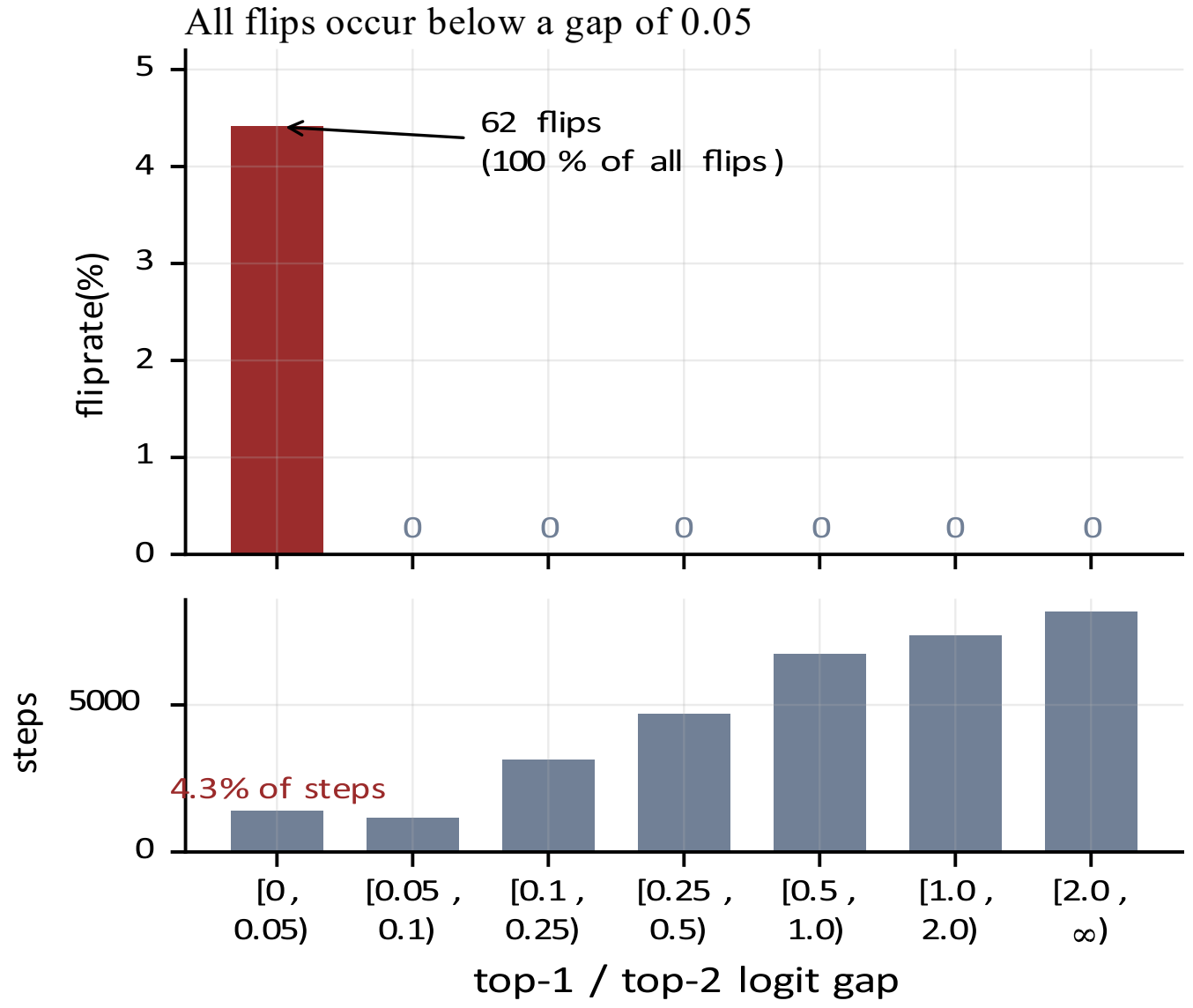

Fig. 1. All observed flips fall in the lowest gap bin, which contains 4.3% of decision steps. Lower panel shows the step count per bin.

The distribution of risk is stronger than "concentrated at low margins". *Every* flip occurred in the lowest gap bin, which accounted for 4.3% of decision steps. No step with a gap above 0.05 flipped, which is consistent with the largest observed perturbation of 0.086.

### C. The perturbation scale is constant

We tested nine candidate predictors of the perturbation's typical size $\sigma$, one at a time, spanning execution configuration, sequence structure, model state and output uncertainty. Four are mentioned here; the remainder appear in Table II.

- Batch size ($B \in \{2,4,8,16,32,64\}$): fitted exponent −0.009; total variation 3% across a 32× range.
- Sequence position: ≈ 5% variation.
- Activation magnitude $\|h\|$: ratio 0.72×, in the direction *opposite* to that predicted by relative floating-point error.
- Top-1 logit magnitude: ratio 1.23×, contradicting the activation result in sign.

A further four predictors, token index during generation, attention entropy, full-distribution entropy and near-tie count – were tested subsequently and are reported alongside the first four in Table II.

TABLE II

ALL NINE TESTED PREDICTORS OF THE PERTURBATION SCALE. RATIO IS $\sigma$ IN THE HIGHEST DECILE OF THE PREDICTOR DIVIDED BY $\sigma$ IN THE LOWEST. QWEN2.5-0.5B; THE LAST FOUR ROWS USE EAGER ATTENTION, UNDER WHICH $\sigma = 0.0753$ AGAINST 0.0837 FOR SDPA.

| Category Predictor | Ratio | Spearman $\rho$ |
|---|---|---|
| batch size | 0.97 | |
| position (prefill) | 1.05 | |
| token index (decode) | 1.18 | +0.( |
| activation norm | 0.72 | −0.( |
| top-1 logit magnitude | 1.23 | |
| attention entropy | 0.94 | +0.( |
| distribution entropy | 0.74 | −0.( |
| near-tie count | 0.81 | −0.( |
| Numeric precision (bf16/fp16) al | 8.0 0 | – |

TABLE III

PRECISION SWEEP. QWEN2.5-0.5B ON NVIDIA L4, 32,640 PREFILL STEPS, $B$=1 VS. $B$=32. IDENTITY CONTROL: 0 FLIPS AT EVERY PRECISION.

| dtype | $\sigma$ | $Q_{0.999}$ | grid step | zeros | flips |
|---|---|---|---|---|---|
| fp32 | 0.0000121 | 0.000055 | 0.0000010 | 94.8% | 0 |
| fp16 | 0.0105424 | 0.046875 | 0.0078125 | 50.5% | 45 |
| bf16 | 0.0837457 | 0.375000 | 0.0625000 | 50.1% | 320 |

No effect exceeds 1.3×. The three predictors that do correlate – activation norm, distribution entropy and near-tie count – correlate with *each other*: uncertain steps have higher entropy, more near-ties and, in this model, smaller activation norms. They are proxies for model confidence rather than for anything mechanical in the arithmetic. The correlation is also *negative*, meaning uncertain steps carry *smaller* perturbations; a per-step margin derived from these signals would widen the margin where the model is confident, which is where flips never occur. Attention entropy is the cleanest null: ratio 0.94, $\rho = +0.006$, $p = 0.39$ despite 24,480 samples. It was the most physically motivated candidate, since diffuse attention sums more terms per output.

We conclude that $\sigma$ is constant for a given model, precision and hardware, and that no conditional noise model is warranted. Only numerical precision moves the perturbation, and it does so exactly (Section III-D). We did not test interactions between axes.

Another point worth noting about the batch sweep experiment is: $B \in \{16,32,64\}$ resulted in *exactly the same* statistics in terms of standard deviation, maximum, percentage of zeros, and the number of flips. This indicates that there is only one reduction rule picked for all three cases of $\delta$, and confirms the interpretation of $\delta$ as finite paths.

### D. Exact scaling with numerical precision

Precision is the one axis on which the perturbation moves, and it moves exactly. Table III reports all three precisions on one model and one GPU.

Perturbation scales exactly with representable precision

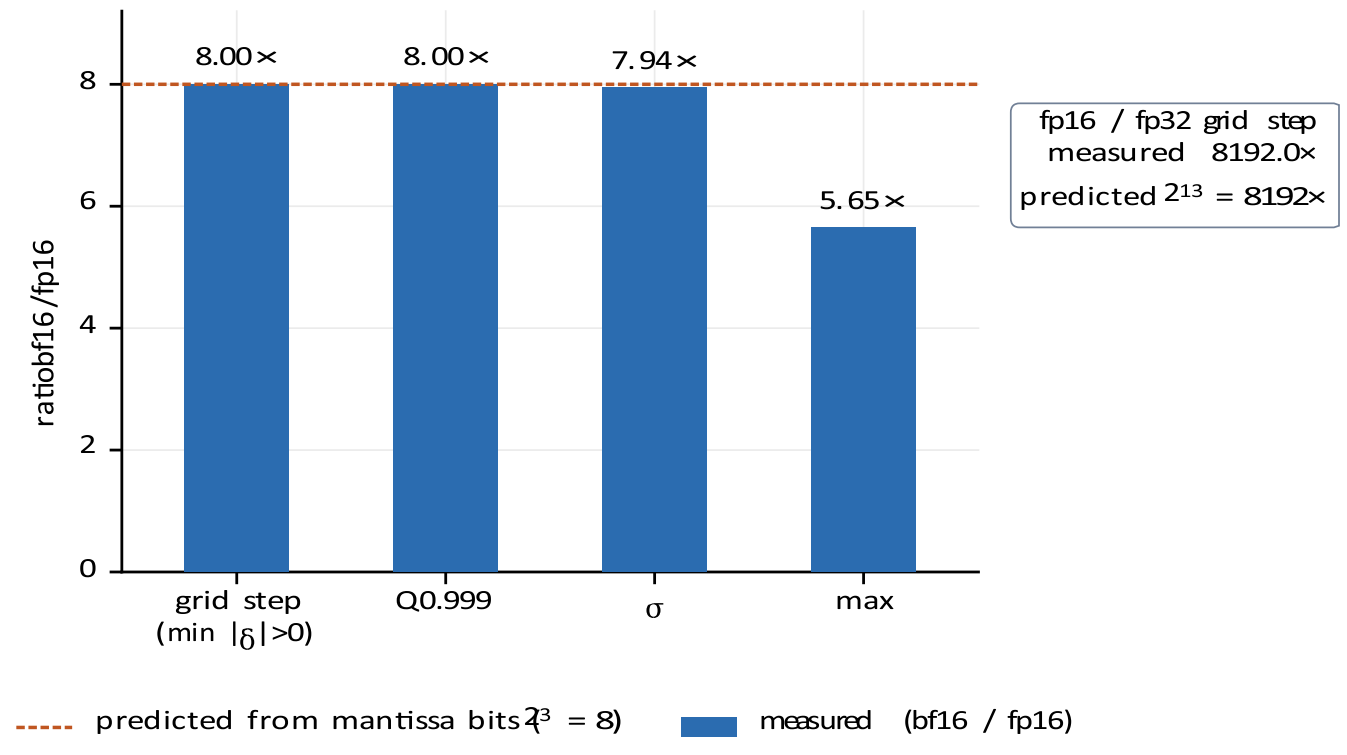

Fig. 2. Measured bf16/fp16 ratios against the mantissa-bit prediction. The *max* column is a single extreme observation rather than a distributional statistic, and is not expected to match.

The *grid step* column is the smallest non-zero perturbation observed, which is the unit in the last place (ulp) at the operative logit magnitude, measured directly with no distributional assumption. Its ratios are 8.00× (bf16/fp16) and 8192.0× (fp16/fp32), against predictions of $2^3$ and $2^{13}$ from the mantissa-bit counts 7, 10 and 23. $Q_{0.999}$ gives the same 8.00×. Figure 2 plots the comparison.

Normalising by each format's own ulp makes the invariance explicit: $\sigma$/ulp is 1.35 and 1.34 for fp16 and bf16, and $Q_{0.999}$/ulp is exactly 6.0 for both. The perturbation distribution is identical across precisions once expressed in grid steps.

Two consequences follow. First, roughly half of all perturbations are exactly zero in fp16 and bf16 because the underlying arithmetic error is smaller than one grid step and rounds away; the measured noise is the true noise seen through a coarse instrument. Second, fp32 departs from the pattern ($Q_{0.999}$/ulp = 55, only 5% zeros) because its grid is fine enough for genuine accumulated error to show through. We call the first regime *quantisation-dominated* and the second *accumulation-dominated*. Deployment precisions sit in the first. Related work on reduced-precision training reports a similar sensitivity of stability to the number format [19].

We also note that fp32 produced zero flips across 32,640 steps. Running in fp32 is a valid determinism strategy at roughly 2× memory and compute, comparable to batchinvariant kernels.

### E. Accidental batch invariance

Batch invariance is not only something a library can be engineered to provide. It also occurs by accident. Table IV sweeps sequence length against batch size for one model, reporting the fraction of decision steps whose gap perturbation is non-zero.

At sequence length 512 the model is batch-invariant at every batch size tested; at 320 and 384 it is invariant at $B$=32 but not at $B$=8 or $B$=16. A practitioner measuring at 512 would observe perfect determinism and conclude the problem does

TABLE IV

FRACTION OF NON-ZERO GAP PERTURBATIONS BY TENSOR SHAPE. QWEN2.5-0.5B-INSTRUCT, BFLOAT16, NVIDIA L4.

| Sequence length | $B$=8 | $B$=16 | $B$=32 |
|---|---|---|---|
| 256 | 53.8% | 54.8% | 54.8% |
| 288 | 56.2% | 55.9% | 55.2% |
| 320 | 54.1% | 54.3% | 0.0% |
| 384 | 53.5% | 54.3% | 0.0% |
| 512 | 0.0% | 0.0% | 0.0% |

not arise for their deployment. Moving to 384 would restore it.

We stress that this is a warning about measurement, not a mitigation. The shapes at which invariance appears are a property of one kernel library on one architecture for one model, and we have not tested their stability.

### F. Cross-platform observation

The same operation, at the same shapes and precisions where CUDA produces a 0.19% flip rate, yields bit-identical output under Apple Metal. We verified this for the real model forward pass at hidden sizes 896, 2048 and 4096 in both fp16 and bf16.

The claim we draw is narrow: batch invariance for this operator class is an implementation choice rather than an inevitable consequence of floating-point arithmetic.

## IV. METHOD

### A. Objective

We add a hinge penalty on the model's own top-1/top-2 logit gap to the standard cross-entropy objective:

$$L = L_{CE} + \lambda \cdot \max\left(0,\ m - (\ell^{(1)} - \ell^{(2)})\right). \tag{3}$$

Cross-entropy is responsible for correctness; the hinge is responsible for decisiveness. When the gap already exceeds $m$ the penalty is exactly zero, so non-fragile steps receive no gradient from the second term and no explicit step-selection rule is required.

For instruction-tuned models we mask both terms to response tokens, so the model is not penalised for uncertainty over the prompt it was given.

### B. Why the hinge targets the model's own top two

An earlier formulation penalised the gap between the *label* token and the best competing token. These coincide only when the model's prediction already matches the label, which on general corpora it frequently does not. Hinging on the label pushes a rank-3 or rank-7 token upward until it arrives near the incumbent top-1, creating near-ties that did not previously exist. In our measurements this formulation moved the median gap from 1.125 to 0.625 and raised the fragile-step fraction to 35.2% the opposite of the intended effect, and worse the stronger the penalty.

Determinism does not depend on which token is correct, only on whether the model commits decisively to whichever token it selects. Equation 3 encodes exactly that.

### C. Worst-case subtraction rather than noise injection

An alternative is to sample perturbations from the measured distribution and add them to the logits during training. We use the hinge instead, for two reasons.

First, defending against the worst case covers the average case by definition, whereas sampling covers the range only in expectation. Second, and concretely, 49% of measured perturbations are exactly zero (Section III-D); sampling from that distribution therefore delivers no learning signal on roughly half of all training steps. The hinge also removes sampling variance from the gradient.

### D. Calibrating the margin

Let $Q_{0.999}$ denote the 99.9th percentile of $|\delta|$, measured for the target model, precision and hardware by the protocol of Section III-A. A single profiling pass suffices.

Surviving the worst case requires the correct token to remain on top after the perturbation has acted against it in both directions: the leading logit may fall by up to $Q_{0.999}$ while the runner-up rises by the same amount. This gives $\ell_{(1)} - Q_{0.999} > \ell_{(2)} + Q_{0.999} \Longleftrightarrow g > 2Q_{0.999},$ (4) so we set $m = 2Q_{0.999}$. The same factor of two appears in prior work [3] from the same argument. Section VI-B showed that the training efficiency empirically peaks at this value.

### E. Profiling transfers from prefill to decode

Profiling is far cheaper in prefill, which yields thousands of decision points per forward pass, than in autoregressive decode, which yields one. The two regimes employ distinct tensor shapes and hence distinct kernels; transfer cannot be assumed beforehand, and the divergence associated with the KVcache decode regime has been documented separately [20].

We computed $Q_{0.999}$ under both regimes using the same model and got 0.2500 under each. Therefore, the calibrated margin is the right target for the decode regime, making the profiling process feasible.

## V. Experimental Setup

### A. Models and hardware

Characterisation covers Qwen2.5 (0.5B, 1.5B, 7B), SmolLM2 (135M, 360M, 1.7B), OPT (125M, 350m), GPT-2 (base, medium, large, distilled), GPT-Neo-125M, Pythia (70M, 160M, 410M, 1B, 1.4B), BLOOM-560M, TinyLlama1.1B and Mistral-7B. Training experiments must be performed with models of size below 0.5B due to the fact that full fine-tuning with float32 master weights and AdamW needs 16 bytes per parameter while our GPU has 22GB of memory. All experiments performed on CUDA were run on NVIDIA L4 except otherwise noted; the previous runs used an NVIDIA Tesla T4 which does not have natively supported bfloat16 and implicitly uses batch-invariant code, which we note since it yielded a negative result. Apple Metal measurements use an M-series Mac under PyTorch MPS.

### B. Data

Characterisation uses WikiText-2. Training and evaluation for the systems experiments use GSM8K [21], with MBPP [22] as an out-of-domain check. Cross-corpus transfer is additionally tested on CodeParrot.

### C. Training configuration

Unless stated: 200 optimiser steps, effective batch 16, learning rate $1\times10^{-5}$ with linear decay to 0.1×, gradient clipping at 1.0, $\lambda = 1$, float32 master weights with bfloat16 autocast, gradient checkpointing enabled. Evaluation is performed in bfloat16, the deployment precision.

### D. Metrics

Fragile-step fraction is the proportion of decision steps whose gap falls below $Q_{0.999}$. It is our primary metric because it is measured over tens of thousands of steps and its confidence intervals are correspondingly tight.

Flip rate is the proportion of steps whose argmax changes under batching. It is the quantity of interest but yields only $10^2$ events per experiment, so differences of 20–30% are not individually significant; we report it as supporting evidence.

Sequence-level divergence is the proportion of generated sequences that differ in any token between $B$=1 and $B$=32. This is what a user experiences and what prior work reports.

Verifier trigger rate at threshold $\tau$ is the proportion of steps with $g < \tau$, i.e. the fraction at which a MarginGatestyle verifier would fire. $\tau_{100}$ denotes the smallest threshold at which every evaluated sequence is deterministic.

Task accuracy is GSM8K exact match on the full 1,319question test set, extracted as the final numeric token of the generated answer.

### E. Statistical treatment

Confidence intervals on rates are bootstrapped over *sequences* or *questions* rather than tokens, because tokens within a sequence are not independent. Paired comparisons between training arms use McNemar's exact test on the per-item outcomes, since the same items pass through both arms. An earlier analysis of the same data treated the arms as independent and overstated confidence; the paired analysis is reported throughout.

## VI. Results

### A. Supervised fine-tuning is a strong baseline

Before evaluating the margin term we note that ordinary in-domain fine-tuning, with no margin term at all, reduced flips by 22% while improving perplexity by 26%. Training on in-domain text makes a model more confident, which widens gaps, which reduces flips - at no additional cost. This is consistent with reports that instruction tuning improves prediction consistency as a side effect [23], although that work varies inputs rather than execution configuration.

All comparisons in this section are therefore against supervised fine-tuning, not against the base model. Comparing against the base model would attribute this free improvement to the method.

TABLE V

MARGIN FRONTIER. QWEN2.5-0.5B, 65,280 EVALUATION STEPS, TWO SEEDS PER POINT, SEED-AVERAGED. $Q_{0.999} = 0.375$ FOR THIS MODEL.

| $m$ | Ppl. cost | Fragility red. | Efficiency |
|---|---|---|---|
| 0.1875 | 0.8% | 2.5% | 3.30 |
| 0.375 | 1.1% | 7.6% | 6.85 |
| 0.75 | 2.2% | 20.5% | 9.39 |
| 1.5 | 7.5% | 38.6% | 5.15 |
| 3.0 | 39.5% | 63.3% | 1.61 |

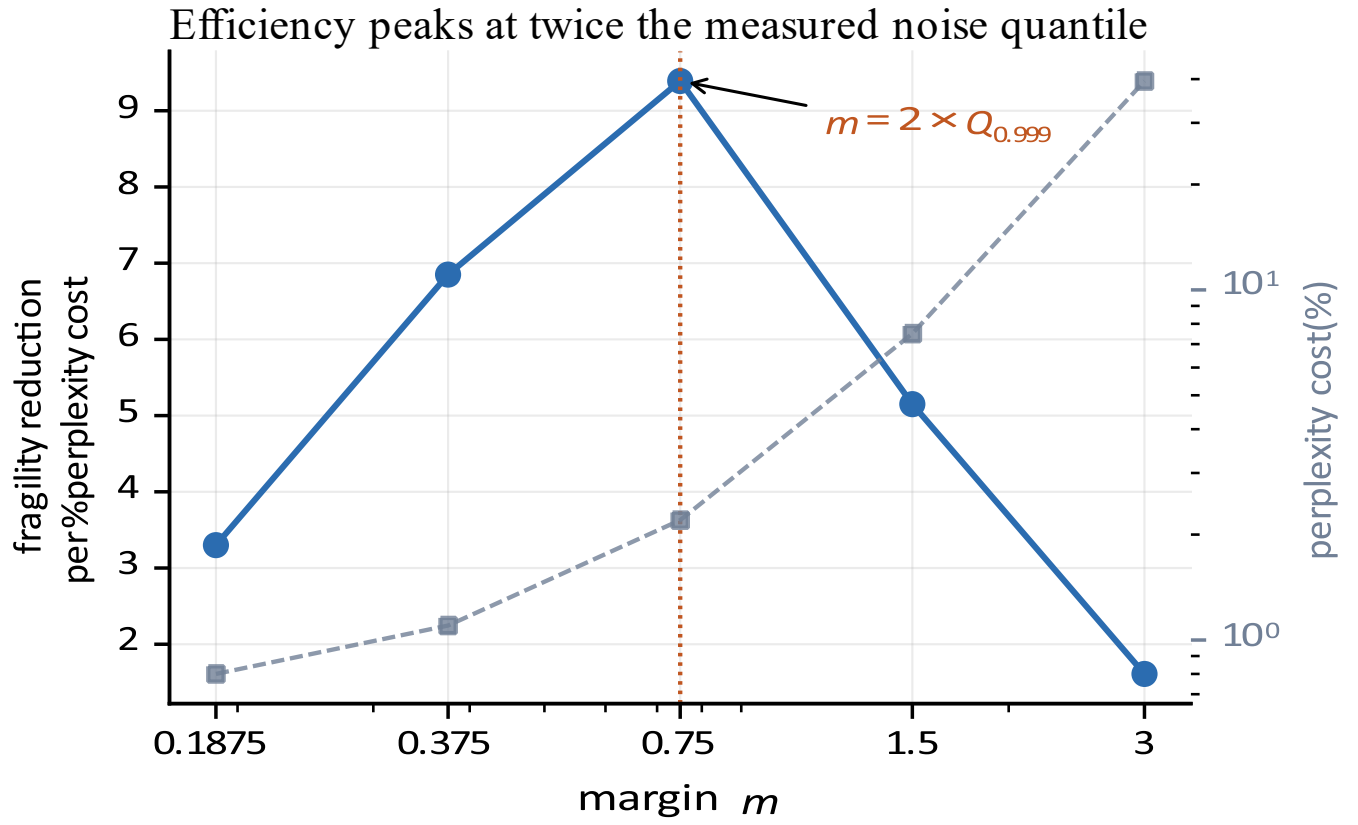

Fig. 3. Fragility reduction per unit perplexity cost, against margin. The peak coincides with $2Q_{0.999}$. Dashed line: perplexity cost, log scale.

## B. Efficiency peaks at the derived margin

Table V sweeps the margin over a 16× range at two seeds per point. We define efficiency as fragility reduction per unit of perplexity cost, both relative to the supervised baseline.

The efficient frontier has an internal maximum at $m = 0.75 = 2Q_{0.999}$, the value that was found in Section. IV. The seed spread was 0.02–0.27 percent points for between margin spreads of 1–5 percent, hence this curve is not noise but signal.It is plotted in Figure. 3.

## C. The peak location transfers across models

Setting margins as multiples of each model's own $Q_{0.999}$ makes the prediction model-independent: if the rule holds, every model should peak at ratio 2 regardless of its absolute margin. Table VI reports six models with predictions registered before training.

Four models in three families peak at ratio 2 with absolute optimal margins ranging from 0.5 to 1.0. Two do not benefit. Under a null in which peak location is uniform over the four tested ratios, all six predictions being correct has probability ≈ 0.002. Section VII reports what happened when this diagnostic was tested further.

## D. Systems evaluation

We evaluate on Qwen2.5-0.5B-Instruct fine-tuned on GSM8K, with loss and hinge masked to response tokens, $m = 2Q_{0.999} = 1.5$. Divergence and verifier statistics use 256 test questions with free-running greedy generation; accuracy uses the full test set.

TABLE VI

CROSS-MODEL VALIDATION. PREDICTIONS WERE PRINTED BEFORE TRAINING. $Q/g$ IS $Q_{0.999}$ DIVIDED BY THE MODEL'S MEDIAN GAP.

| Model | Family | $Q/g$ | Pred. | Peak |
|---|---|---|---|---|
| OPT-350M | OPT | 0.29 | 2.0 | 2.0 |
| SmolLM2-135M | Llama | 0.38 | 2.0 | 2.0 |
| Qwen2.5-0.5B | Qwen | 0.40 | 2.0 | 2.0 |
| SmolLM2-360M | Llama | 0.50 | 2.0 | 2.0 |
| GPT-2-medium | GPT-2 | 1.00 | fail | 4.0 |
| Pythia-410M | NeoX | 3.13 | fail | none |

TABLE VII

SYSTEMS EVALUATION ON GSM8K. PAIRED MCNEMAR TESTS; ACCURACY ON $n$=1,319, DIVERGENCE ON $n$=256.

| | SFT | Margin | $p$ |
|---|---|---|---|
| Sequences diverged | 53.8% | 38.1% | 0.004 |
| Mismatched steps | 197 | 128 | – |
| GSM8K accuracy | 35.94% | 33.06% | 0.023 |
| Mean output length | 119 | 121 | – |

Sequence-level divergence falls from 53.8% to 38.1%, a 29% relative reduction, at matched output length. Length matching matters: a shorter answer has fewer opportunities to diverge, so an unmatched comparison would be uninformative.

The verifiers' trigger rate drops by 14–19% when thresholds are below $\tau = 2$, with non-zero bootstrap confidence intervals for the difference. At $\tau = 4$ the difference becomes insignificant, while at $\tau = 8$ the drop turns into an increase – we discuss this issue in Section VII.

## E. The accuracy cost

Table VIII reports that the margin sweep against task accuracy on the full test set.

Accuracy decreases monotonically with respect to margin and becomes statistically significant at $m = 1.0Q$. This is the trade-off that had been predicted in previous studies of entropy collapse in heavily tuned models [16] and post-instruction tuning overconfidence [17]. The prior result with $n = 160$, which found no penalty, does not hold up to the larger dataset.

The relationship between accuracy and verifier load is depicted in Figure 4. Accuracy falls cleanly; verifier load does not, for reasons given in Section VII.

# VII. WHAT THE METHOD CANNOT DO

## A. The worst case does not move

The main result is that margin training decreases the frequency of flips but not the worst-case size of gap at which flips can happen. Table IX gives the distribution of gap sizes at which the remaining flips have happened.

Flip frequency falls monotonically, from 197 to 128. The gaps at which the remaining flips occur do not change: the median is 0.125 in every arm and the 95th percentile is flat. Margin training removes flips; it does not make the survivors harder to produce.

TABLE VIII
ACCURACY AGAINST MARGIN, GSM8K FULL TEST SET ($n$ = 1,319), PAIRED MCNEMAR AGAINST THE SUPERVISED BASELINE.

| Arm | $m$ | Accuracy | Difference | $p$ |
|---|---|---|---|---|
| SFT | — | 35.94% | — | — |
| 0.5$Q$ | 0.375 | 34.65% | −1.29 | 0.326 |
| 1.0$Q$ | 0.750 | 33.51% | −2.43 | 0.058 |
| 2.0$Q$ | 1.500 | 33.06% | −2.88 | 0.023 |

Accuracy falls monotonically; verifier load does not

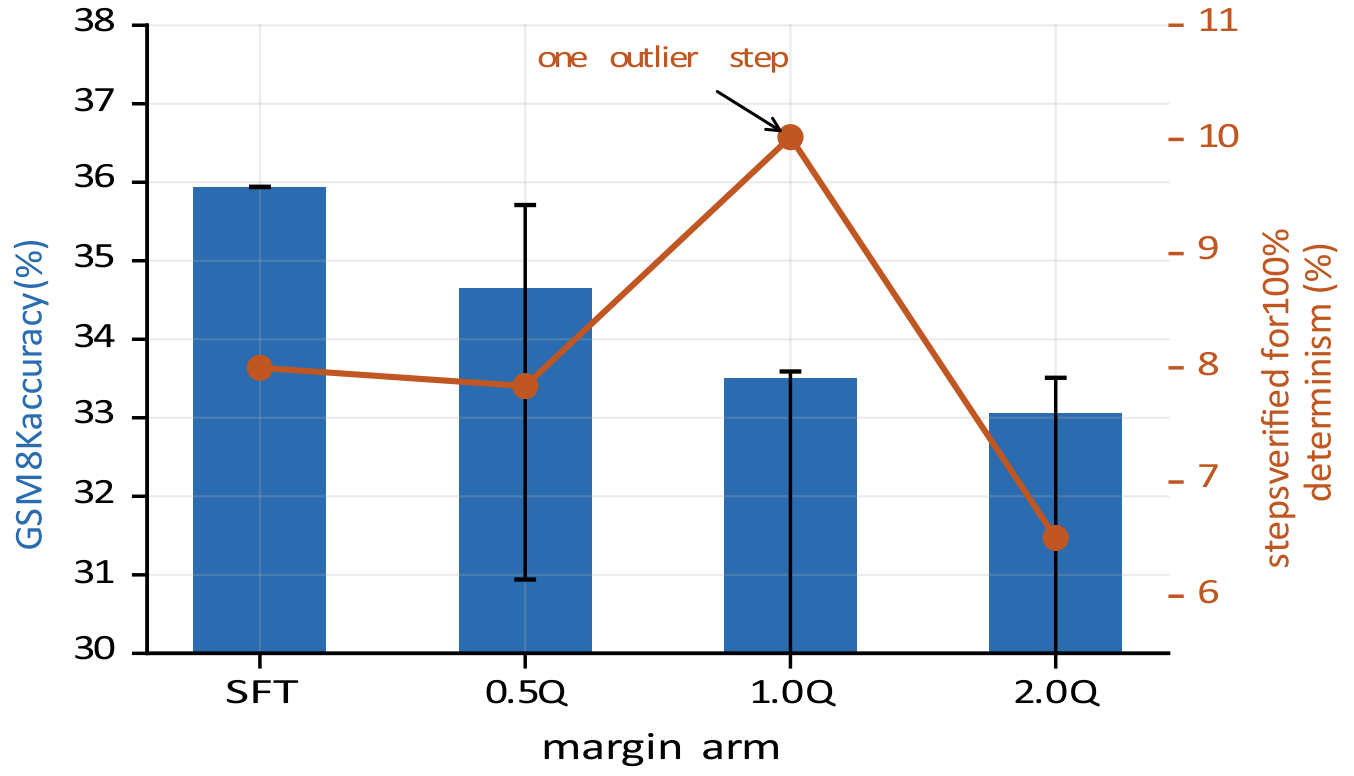

Fig. 4. GSM8K accuracy (bars, with 95% paired bootstrap intervals) against steps requiring verification for full sequence determinism (line).

*a) Why.:* Two things are involved here and neither is removable via tuning. First, the hinge is a soft constraint and not a hard one, since it imposes a cost in case of a small gap, but does not preclude the latter. Second, requiring that *all* gaps be above the noise ceiling is asking for something the model cannot achieve since the increase in perplexity cost will exceed the gain, getting to be 39.5% at $m$ = 3.0 for one model and 1679% on another. However, $Q_{0.999}$ remains equal to 0.75 before and after training: the effect of hardware perturbation remains the same.

Frequency is trainable and the worst case is dictated by hardware.

### B. Consequence: $\tau_{100}$ does not improve

Since $\tau_{100}$ depends on the worst mismatched step among all the considered sequences, this becomes its property too. This can be seen in Table X.

The 1.0$Q$ arm had one flip at gap 0.875, requiring $\tau_{100}$ to jump to the next grid point and thereby increasing its verifier cost by one quarter. One step decides the column, so we make no claim about a decrease in the number of verifiers invoked for fully deterministic outcomes; the 18.5% improvement in 2.0$Q$ is no more reliable than the 25% degradation in 1.0$Q$.

### C. The margin loss redistributes rather than widens

The far-right column of Table IX indicates the impact of the loss on the gap distribution: the $5^{th}$ percentile increases by 33%, whereas the $95^{th}$ percentile falls by 14%. The distribution *compresses* (Figure 5). This explains the triggerrate reversal at $\tau$ = 8: with $m$ = 1.5 the hinge has no reason

TABLE IX
GAPS AT WHICH SURVIVING FLIPS OCCUR, AND THE FULL GAP DISTRIBUTION. QWEN2.5-0.5B-INSTRUCT ON GSM8K, ≈30,000 RESPONSE-TOKEN STEPS.

| | | gap at flip | | | all steps | | |
|---|---|---|---|---|---|---|---|
| Arm | Flips | max | p95 | med | p5 | p50 | p95 |
| SFT | 197 | 0.500 | 0.250 | 0.125 | 0.375 | 6.250 | 13.875 |
| 0.5$Q$ | 177 | 0.625 | 0.250 | 0.125 | 0.375 | 6.000 | 12.875 |
| 1.0$Q$ | 139 | 0.875 | 0.250 | 0.125 | 0.500 | 5.875 | 12.438 |
| 2.0$Q$ | 128 | 0.625 | 0.206 | 0.125 | 0.500 | 6.000 | 12.000 |

TABLE X
VERIFIER INVOCATIONS REQUIRED FOR 100% SEQUENCE DETERMINISM.

| Arm | $\tau_{100}$ | Steps verified | vs. SFT |
|---|---|---|---|
| SFT | 0.75 | 8.00% | – |
| 0.5$Q$ | 0.75 | 7.84% | +2.0% |
| 1.0$Q$ | 1.00 | 10.02% | −25.2% |
| 2.0$Q$ | 0.75 | 6.51% | +18.5% |

to preserve gaps at 5 or 6, and cross-entropy trades some of them away.

The benefit is therefore threshold-dependent: helpful below $\tau \approx 2$, neutral at 4, mildly harmful at 8.

### D. A viability diagnostic that failed

Section VI reported that four models with $Q/g < 1$ peaked at ratio 2 while two with $Q/g \geq 1$ did not, under pre-registered prediction. This suggested a diagnostic computable before any training.

We tested it by searching for models in the unexplored band $Q/g \in [0.55, 0.95]$. Two were found, both at *exactly* 0.67:

| Model | $Q/g$ | zeros | Peak |
|---|---|---|---|
| GPT-Neo-125M | 0.67 | 3.6% | 2.0 (holds) |
| DistilGPT-2 | 0.67 | 64.5% | 4.0 (fails) |

Identical diagnostic value, opposite outcomes. The diagnostic is not sufficient. Nor does the zero fraction rescue it: GPTNeo-125M at 3.6% zeros is more accumulation-dominated than Pythia-410M (13.9%, which fails) and yet it works.

The only surviving pattern is that all three failures are of GPT-2 lineage or Pythia, and both GPT-2 failures show negative fragility reduction at small margins – the penalty makes them worse before it helps. No model that works does this. We cannot convert this into a usable predictor, and we report it as an open question rather than a contribution.

### E. Benefits are in-domain only

Trained on WikiText and evaluated elsewhere, the fragility reduction is 0.87× in domain but 0.95× on GSM8K and 0.97× on code. The median gap on the out-of-domain corpora is unchanged. Trained on GSM8K and evaluated on MBPP, every verifier triggerratio lies between 0.97 and 1.02 and divergence is slightly worse.

The margin loss compresses the gap distribution

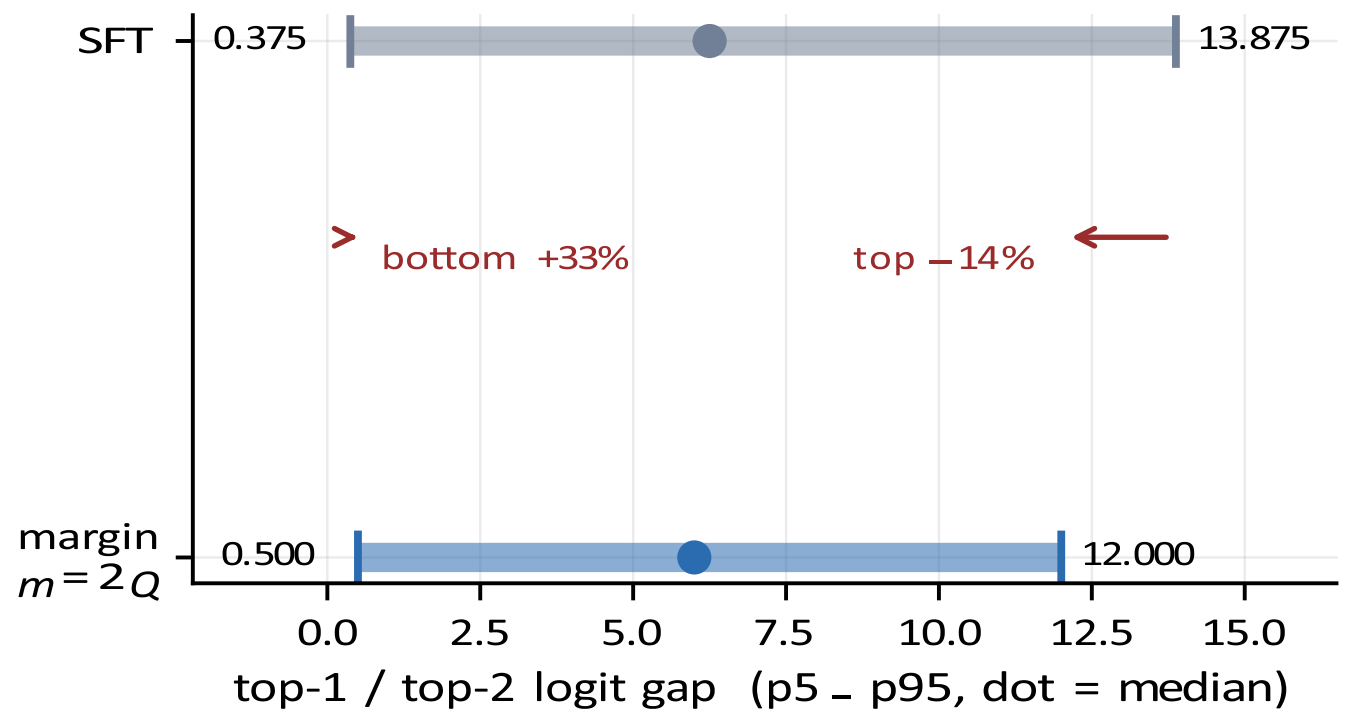

Fig. 5. The gap distribution before and after margin training (5th–95th percentile, dot at the median). The bottom rises and the top falls.

The model learns to be decisive about the fine-tuning distribution, not decisive in general. Any deployment claim must be scoped to the distribution the model was trained on.

### F. Evaluation setting dominates model scale

An early attempt to measure verifier trigger rates found no reduction at any threshold. The cause was the evaluation setting, not the method: a base model predicting unseen prose is uncertain everywhere, giving trigger rates near 88% against the 18.56% reported for instruction-tuned models on reasoning benchmarks [3]. A margin widening of 0.1 logits is invisible against a threshold of 4.0.

We verified that this is not a scale effect: 7B base models measured at 80.8% and 78.7% trigger rate at $\tau = 4$, essentially the same as the 350M model's 87.6%. Switching to an instruction-tuned model on a real task, measured on response tokens only, moved the trigger rate to 40.9% and produced the reductions reported in Section VI.

This is a methodological warning for the field: flip and trigger statistics are not comparable across evaluation settings, and base-model-on-general-corpus measurements will understate the value of any gap-widening intervention.

## VIII. Limitations

*a) Model scale:* All training experiments use models below 0.5B parameters, because full fine-tuning with float32 master weights requires 16 bytes per parameter and our hardware provides 22GB. We verified only that the characterisation transfers: at 7B, $Q/g$ is 0.40 (Qwen2.5-7B) and 0.19 (Mistral7B-v0.3), with flip rates of 1.04% and 0.42% inside the range reported by prior work. These measurements use 4-bit weight loading, which alters the arithmetic, and are indicative rather than definitive. Whether the margin frontier retains its shape at 7B is untested.

*b) Single hardware family:* All CUDA results come from two NVIDIA architectures (Turing, Ada). We found that $\sigma$ is nearly identically transferred between them (0.01062 vs. 0.0105424 at float16 precision) but the maximum is not (0.0859 vs. 0.1328), in line with $\sigma$ as a bulk quantity and the maximum as a path-dependent quantity. We have not tried Hopper or Blackwell, or any accelerator not produced by NVIDIA besides Metal.

*c) Characterisation is prefill-dominant:* Most measurements use prefill because it yields three orders of magnitude more decision points per unit of compute. We verified that $Q_{0.999}$ is identical in both regimes and that the fragility reduction transfers (0.87× prefill, 0.80× decode), but the decode flip-count comparison was underpowered at our sample size.

*d) Single task for the systems evaluation:* The systems results are GSM8K only. MBPP served as an out-of-domain check and showed no transfer, which is itself reported, but we have not established the method's behaviour on a second in-domain reasoning task.

*e) Unexplained model-dependence:* We do not know which architectures will benefit. Unexplored hypotheses for explanation include the size of the activation at unembedding, different architectures like parallel attention and feedforward networks, and pretraining differences. This is what we believe to be the most important open problem posed by our research.

*f) Margin and hinge weight not jointly tuned:* The sweeps of $m$ were done at a fixed value of $\lambda = 1$, while the lack of an influence of $\lambda = 0.1$ was verified. The joint space $(\lambda, m)$ remains unstudied, and another combination of parameters can affect the accuracy penalty.

*g) Short training schedules:* Each arm will use 200 optimisation steps. The test is whether the order on the Pareto front holds up with further training.

## IX. Conclusion

Our aim was to see if a language model can be trained to withstand the effect of floating point noise introduced via batching such that the cost of being deterministic is cheaper. The response to this is an affirmative one, with clear boundaries.

The use of margin regularisation at a calibrated margin leads to a reduction in token flips of 35% and sequence divergence of 29%, but at the cost of 1.3-2.9 accuracy that grows monotonically with the margin. The efficiency-optimal margin is twice the measured noise quantile, a value derived rather than tuned, and it transfers across four model families.

It does not, and cannot, deliver determinism. The hinge is a soft penalty and the hardware's largest perturbation is unchanged by training, so the worst-case gap at which a flip can occur stays where it was. Margin training and worstcase verification therefore do not compose: an inference-time verifier still needs the same threshold, and only fires slightly less often.

Two measurement results stand independently of the method. The perturbation scales exactly with representable precision, so a margin measured at one precision can be converted to another by mantissa-bit arithmetic. And batch invariance arises accidentally at particular tensor shapes, which means any published flip rate is uninterpretable without its sequence length and batch size – a practitioner can measure zero nondeterminism simply by choosing a sequence length of 512.

We also report seven rejected hypotheses, including a viability diagnostic that survived pre-registered prediction on six models before being refuted by the seventh. They are included because they are what give permission to the final design, that is, the one measured constant, since four alternative possibilities have been tested and found to be wrong.

The most useful open question is why some model families do not benefit. A predictor for that would make a technique that fails inexplicably into one that can be used with confidence.

## REPRODUCIBILITY

The measurement code, the entire experiment log containing failed attempts, and the per-attempt raw results can be found at GitHub